\documentclass{article} % For LaTeX2e
\usepackage{iclr2026_conference,times}
\usepackage{graphicx}
\usepackage{enumitem}

\definecolor{lightgray}{gray}{0.5}
\usepackage{multirow}
\usepackage{booktabs}
\usepackage{colortbl}
\usepackage{bbm}
\usepackage{subcaption}
\usepackage{wrapfig}
\usepackage{float}

\usepackage{amsmath,amsfonts,bm}

\def\eqref#1{equation~\ref{#1}}
\def\1{\bm{1}}

\def\vmu{{\bm{\mu}}}

\def\vc{{\bm{c}}}
\def\vd{{\bm{d}}}

\def\vg{{\bm{g}}}

\def\vo{{\bm{o}}}

\def\vr{{\bm{r}}}
\def\vs{{\bm{s}}}

\def\vu{{\bm{u}}}

\def\mK{{\bm{K}}}

\def\mP{{\bm{P}}}

\def\mT{{\bm{T}}}

\DeclareMathAlphabet{\mathsfit}{\encodingdefault}{\sfdefault}{m}{sl}
\SetMathAlphabet{\mathsfit}{bold}{\encodingdefault}{\sfdefault}{bx}{n}

\newcommand{\R}{\mathbb{R}}

\newcommand{\vC}{\boldsymbol{C}}

\newcommand{\vF}{\boldsymbol{F}}

\newcommand{\vI}{\boldsymbol{I}}

\newcommand{\vM}{\boldsymbol{M}}

\usepackage{hyperref}
\usepackage{url}

\iclrfinalcopy

\title{AESplat: Advancing Pose-Free Feed-Forward 3D Gaussian Splatting via Decoupled Appearance Modeling}

\author{Shiwei Ren, Zhiang Liu, Yongchun Fang\thanks{Corresponding author.}, Hongwei Chen \\
Nankai University \quad renshiwei@mail.nankai.edu.cn\\
\url{https://github.com/aesplat/AESplat}
}

\begin{document}

\maketitle

\begin{figure}[!h]
    \centering 
    \includegraphics[width=\textwidth]{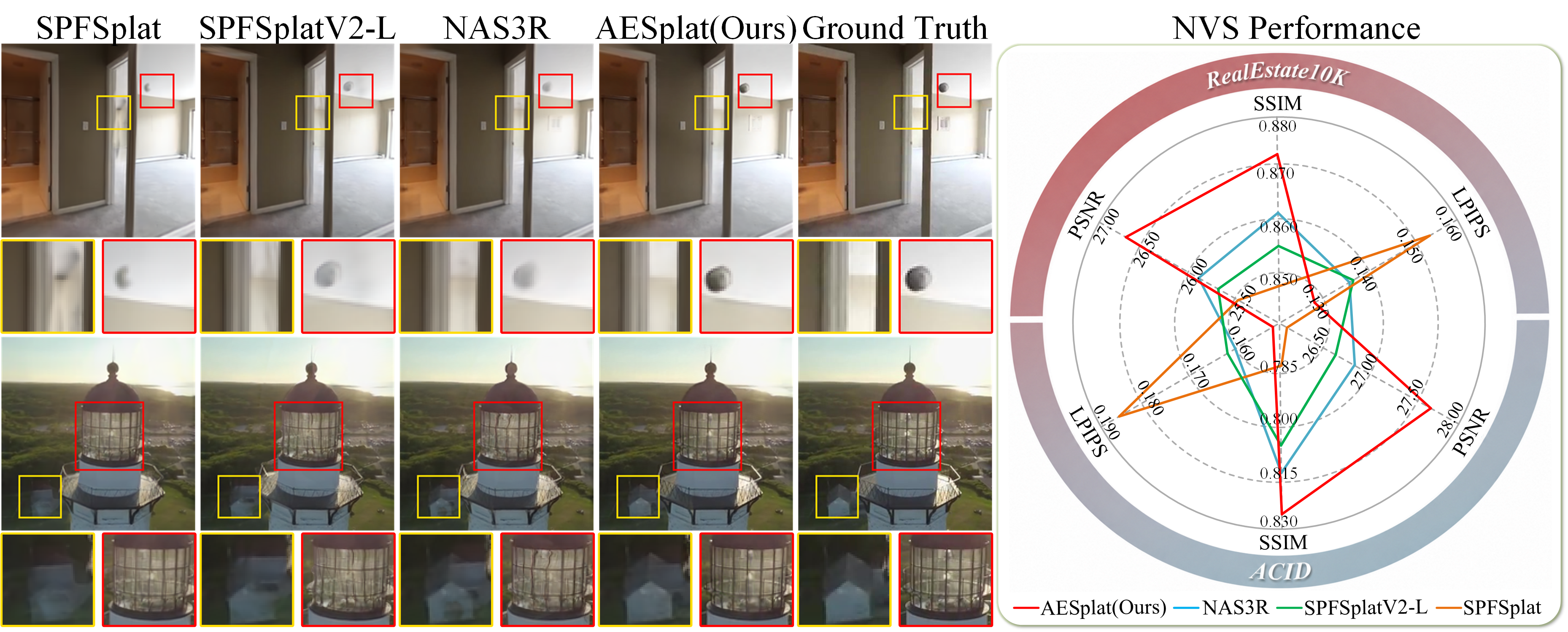} 
    \caption{Our method AESplat consistently outperforms state-of-the-art pose-free feed-forward methods in rendering quality across both indoor and outdoor scenes. Yellow and red boxes highlight differences in diffuse (e.g., walls) and specular (e.g., mirrors and glass) appearance, respectively.}
    \label{fig:teaser}
\end{figure}
% \begin{figure}[!h]
%     \centering 
%     \includegraphics[width=\textwidth]{figures/teaser.png} 
%     \caption{Our method AESplat consistently outperforms state-of-the-art feed-forward methods (DepthSplat~\citep{xu2025depthsplat}, SPFSplat~\citep{huang2025no}, and NAS3R~\citep{huang2026none}) in rendering quality across both indoor and outdoor scenes. The scenes are drawn from the RE10K~\citep{10.1145/3197517.3201323} dataset (top row) and ACID~\citep{liu2021infinite} dataset (bottom row). View-independent appearance differences on diffuse surfaces (e.g., walls) are highlighted in yellow, while view-dependent effects on specular surfaces (e.g., mirrors and glass) are highlighted in red.}
%     \label{fig:teaser}
% \end{figure}
\begin{abstract}
Pose-free feed-forward 3D Gaussian Splatting (3DGS) has demonstrated remarkable potential for generalized novel view synthesis. However, existing methods typically predict Gaussian appearance attributes represented by spherical harmonics (SH) in the same manner, overlooking the fundamental distinction between view-independent and view-dependent appearance, which results in suboptimal rendering quality. In this paper, we present AESplat, a novel and general framework for pose-free feed-forward 3DGS that introduces an effective decoupled appearance modeling strategy based on an analysis of SH, enabling higher-quality rendering. Specifically, AESplat directly derives the zeroth-order SH coefficient, which represents the base view-independent appearance component, from the input images without training. The higher-order SH coefficients are subsequently predicted by a shallow multilayer perceptron equipped with two efficient 3D-aware inductive biases to model view-dependent appearance variations. Extensive experiments across multiple datasets demonstrate that our method significantly outperforms state-of-the-art approaches, achieving a $0.8$ dB improvement in PSNR over the pose-free method NAS3R and a $1.1$ dB improvement over the pose-required method DepthSplat on the RealEstate10K dataset.

% Project page: \url{https://aesplat.github.io/}.

\end{abstract}
\section{Introduction}
Novel view synthesis (NVS) is a fundamental problem in computer vision and graphics, with broad applications in virtual reality, autonomous driving, and robotics. 3D Gaussian Splatting (3DGS)~\citep{kerbl20233d} enables high-fidelity real-time rendering, yet conventional approaches rely on costly per-scene optimization, hindering their deployment in user-friendly applications. Feed-forward 3DGS overcomes this limitation by directly regressing Gaussian primitives from input images in a single forward pass, enabling efficient and generalizable NVS for unseen scenes.

Most feed-forward 3DGS methods rely on accurate camera poses from SfM~\citep{schonberger2016structure}, limiting their applicability to unposed image collections. To eliminate this dependency, recent studies have explored pose-free feed-forward 3DGS, reconstructing scenes directly from unposed images. Notably, many pose-free methods build upon powerful 3D foundation models (3DFMs) designed for geometric perception, such as point clouds, depth maps, and camera poses. For instance, several existing methods~\citep{smart2024splatt3r,ye2025no,huang2025no} build on MASt3R~\citep{leroy2024grounding}, while others~\citep{jiang2025anysplat,huang2026none} leverage VGGT~\citep{wang2025vggt}. Powered by their pretrained representations, these pose-free methods achieve high-quality rendering and can even outperform pose-required approaches.

\begin{figure*}[t]
    \centering
    \includegraphics[width=\linewidth]{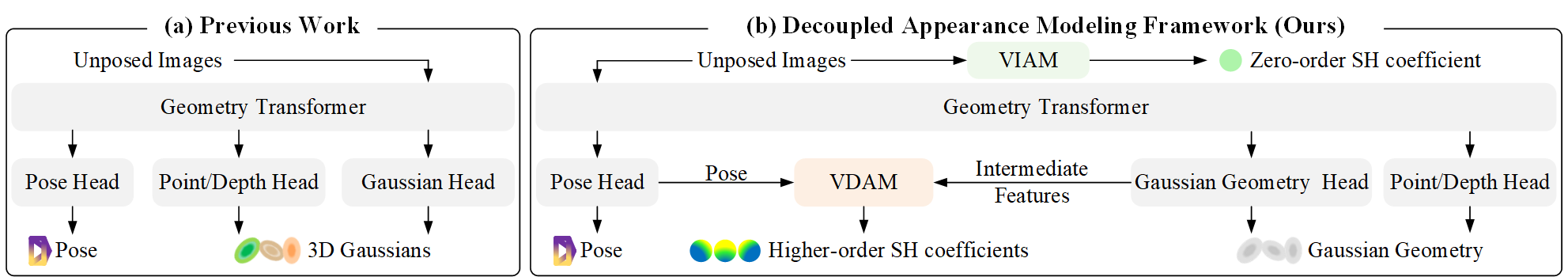}
    
    \caption{\textbf{Architectural Differences.} Unlike (a) existing pose-free methods that predict all appearance attributes with a unified prediction head, (b) we adopt a decoupled formulation that employs tailored strategies to model view-independent and view-dependent appearance separately.}

    \phantomsubcaption\label{fig:difference_a}
    \phantomsubcaption\label{fig:difference_b}

    \label{fig:difference}
\end{figure*}

Despite encouraging progress, pose-free methods still exhibit a substantial rendering-quality gap compared to optimization-based approaches. As illustrated in Fig.~\ref{fig:difference_a}, building on pretrained 3DFMs, recent methods typically use predicted point clouds or depth maps to determine Gaussian positions, while the same multi-view features from the geometry transformer are used to predict the remaining geometric attributes. In contrast, all spherical harmonics (SH) coefficients representing the Gaussian appearance are jointly predicted by an additional Gaussian head. This disparity suggests that geometry modeling increasingly benefits from advances in 3DFMs, whereas appearance modeling remains relatively simple and largely follows a homogeneous prediction paradigm. This motivates us to investigate appearance modeling as an important yet underexplored factor for improving rendering quality. As discussed in Sec.~\ref{method:motivation}, the zeroth-order SH basis function is view-independent, whereas higher-order SH terms are view-dependent, indicating their corresponding coefficients play distinct roles in representing appearance. Therefore, simply predicting these two appearance components in the same manner may limit the model's rendering capability.

To address this limitation, we propose AESplat, a novel and general framework for pose-free feed-forward 3DGS, as illustrated in Fig.~\ref{fig:difference_b}, which decoupledly models appearance to improve rendering quality. Our key insight is that SH coefficients play fundamentally different roles in appearance representation. The main challenge lies in simultaneously and effectively modeling both view-independent and view-dependent appearance. For view-independent appearance, we observe that, under the pixel-aligned Gaussian paradigm, the pixel already provides a strong observation of the base appearance for its corresponding Gaussian. Instead of re-predicting this explicitly observed information, we directly derive the zeroth-order SH coefficient from the pixel RGB value without training. For view-dependent appearance, we believe that explicit 3D-aware inductive biases remain crucial for correctly modeling this multi-view attribute. Based on this, we design two effective inductive biases: Gaussian-to-Views Spatial Relation Embedding (GVSRE) and Warped Color Map (WCM). Furthermore, we use a shallow multilayer perceptron (MLP) that uses GVSRE and WCM as inputs to predict higher-order SH coefficients to effectively model view-dependent appearance. As shown in Fig.~\ref{fig:teaser}, our method effectively models these two appearances, substantially improving rendering quality. In summary, our main contributions are as follows: 

\begin{itemize}[leftmargin=8pt,itemsep=0.2pt, topsep=0.2pt]
    \item We propose AESplat, a novel and general framework for pose-free feed-forward 3DGS, which revisits Gaussian appearance modeling through the lens of SH and shift from unified to decoupled appearance modeling, enabling higher-quality rendering.

    \item We develop an effective decoupled strategy for appearance modeling. The zeroth-order SH coefficient of pixel-aligned Gaussians is directly obtained from the input images in a training-free manner. The higher-order SH coefficients are predicted by a shallow MLP conditioned on two effective 3D-aware inductive biases, GVSRE and WCM.

    \item Extensive experiments on multiple challenging benchmarks demonstrate that our method consistently achieves new state-of-the-art (SOTA) performance in rendering quality among both pose-free and pose-required feed-forward 3DGS methods. 
\end{itemize}

\section{Related Work}
\label{relatedwork}

\noindent\textbf{Pose-required Feed-Forward 3DGS}. Feed-forward 3DGS has emerged as a mainstream paradigm for more generalizable NVS, without per-scene optimization. Most existing feed-forward 3DGS methods~\citep{charatan2024pixelsplat,chen2024mvsplat,chen2024mvsplat360,zhang2025transplat,xu2025depthsplat,xu2025resplat,ye2026yonosplat} rely on accurate camera poses obtained via SfM for differentiable rendering during training and inference. Additionally, some methods~\citep{zhang2024gs,tang2024lgm,ICLR2025_9676c528} convert camera poses into Plücker ray maps~\citep{plucker1865xvii}, which are jointly fed into the transformer with the input images to predict Gaussian representations. However, their reliance on accurate camera poses limits their applicability to unposed image collections. In contrast, our approach operates in the pose-free setting, eliminating the need for ground-truth camera poses.

\noindent\textbf{Pose-free Feed-Forward 3DGS}.
Recent work has increasingly explored pose-free feed-forward 3DGS without ground-truth camera poses. NoPoSplat~\citep{ye2025no} predicts 3D Gaussians in a canonical coordinate system directly from unposed images and further estimates relative camera poses during inference. To remove the need for ground-truth poses during rendering, several self-supervised pose-free frameworks have been proposed. PF3Splat~\citep{hong2025pfplat} estimates camera poses from predicted depth and point correspondences via classical pose estimation techniques, followed by coarse-to-fine Gaussian reconstruction. SPFSplat~\citep{huang2025no} and its successor SPFSplatV2~\citep{huang2025spfsplatv2} unify Gaussian prediction and camera pose estimation within a single feed-forward architecture. More recently, NAS3R~\citep{huang2026none} further demonstrates strong reconstruction capability without pretrained weights. Despite their promising performance, these methods uniformly regress Gaussian appearance without explicitly accounting for the distinct roles of zeroth-order and higher-order SH components in appearance representation, limiting rendering quality. Instead, our method adopts a decoupled appearance modeling framework with dedicated designs for each component to achieve higher-quality rendering.

\noindent\textbf{Appearance Modeling in 3DGS}. 3DGS~\citep{kerbl20233d} typically models Gaussian appearance with SH, while subsequent works improve its expressiveness through shading functions~\citep{jiang2024gaussianshader}, anisotropic spherical Gaussian appearance fields~\citep{NEURIPS2024_708e0d69}, or mirrored counterparts~\citep{liu2024mirrorgaussian}. However, their per-scene optimization limits efficiency and generalization. Recently, some pose-free feed-forward methods~\citep{zhang2025flare,zhao2026rayzer,jeong2026viewsplat} leverage estimated camera poses to either predict all Gaussian attributes or fine-tune them using predicted attribute offsets. However, they still treat all SH coefficients uniformly, without distinguishing their fundamentally different roles. Unlike these methods, we explicitly exploit the functional distinction between zeroth-order and higher-order SH components and adopt tailored modeling strategies for them to improve rendering quality.

\section{BackGround: Pose-free Feed-forward 3D Gaussian Splatting}
\label{bg}

Our method is built upon NAS3R~\citep{huang2026none}, a recent pose-free feed-forward 3D reconstruction method. It jointly estimates camera parameters and reconstructs 3D Gaussians from $N$ unposed input images $\{{\displaystyle \vI}^v\}_{v=1}^N$, which comprise $N_\mathcal{C}$ context images $\mathcal{I}_\mathcal{C}$ and $N_\mathcal{T}$ target images $\mathcal{I}_\mathcal{T}$.

\noindent\textbf{Masked Geometry Transformer.}
The geometry transformer employs a ViT-based encoder-decoder architecture~\citep{dosovitskiyimage} to extract multi-view features. Following the backbone design, each input image is first patchified and encoded by DINOv2~\citep{oquab2023dinov2} into image features ${\displaystyle \vF}^v$. These features are then concatenated with a learnable camera token ${\displaystyle \vg}^v$ to form the input tokens for the decoder. The decoder employs Alternating-Attention~\citep{wang2025vggt} to aggregate information across multiple views, with masked attention~\citep{huang2025spfsplatv2} restricting interactions such that context tokens attend only to context tokens, while target tokens attend to both. This prevents target-view information from affecting Gaussian reconstruction while enabling pose estimation from global scene context. 

\noindent\textbf{Camera Parameter Estimation.}
Given refined camera tokens $\hat{{\displaystyle \vg}}^v$, the camera head predicts camera extrinsics ${\displaystyle \mP}^{v\rightarrow1}$ and intrinsics ${\displaystyle \mK}^v$ for each view. Extrinsics use a 6D rotation representation and a 4D homogeneous translation vector, which are converted into a homogeneous transformation matrix. The first view is defined as the canonical frame. Intrinsics are parameterized by the field of view (FOV), assuming equal horizontal and vertical FOVs and a centered principal point.

\noindent\textbf{Gaussian Prediction.}
Given refined context tokens $\hat{{\displaystyle \vF}^v_\mathcal{C}}$, the Gaussian prediction module estimates 3D Gaussians using two DPT~\citep{ranftl2021vision} heads. One predicts depth to recover Gaussian centers with the estimated camera parameters, while the other predicts rotation, scale, opacity, and SH coefficients. All Gaussians are transformed into the canonical frame of the first view:
\begin{equation}
    \{\mathcal{G}^{v \to 1}\}_{v=1}^{N_\mathcal{C}}
    =
    \{({\displaystyle \vmu}_j^{v\to1}, {\displaystyle \vr}_j^{v\to1}, {\displaystyle \vs}_j, \alpha_j, {\displaystyle \vc}_j^{v\to1})\}_{j=1}^{H\times W},
    \label{eq:gaussianparams}
\end{equation}
where ${\displaystyle \vmu} \in {\displaystyle \R}^3$, ${\displaystyle \vr} \in {\displaystyle \R}^4$, ${\displaystyle \vs} \in {\displaystyle \R}^3$, $\alpha \in {\displaystyle \R}$, and ${\displaystyle \vc} \in {\displaystyle \R}^{(l+1)^2 \times 3}$ denote the Gaussian center, rotation quaternion, scale, opacity, and SH coefficients, respectively, with $l$ denoting the SH degree.
\section{AESplat}
\label{method}

\begin{figure}[t]
    % \vspace{-1.0cm}
    \centering 
    \includegraphics[width=\textwidth]{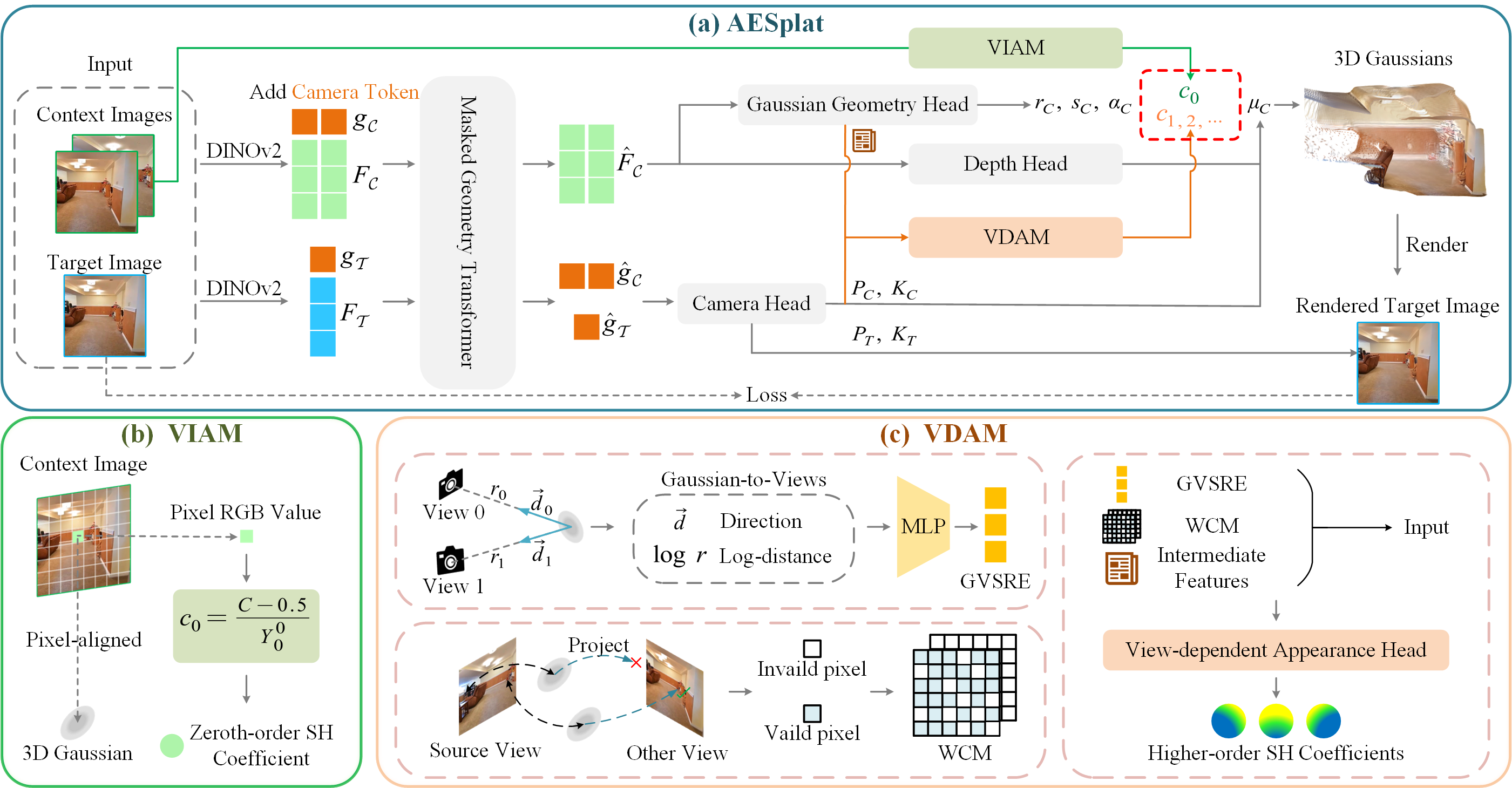} 
    % \vspace{-0.7cm}
    \caption{\textbf{Pipeline of AESplat}. (a) Given unposed images, AESplat jointly predicts camera poses and Gaussian primitives in a single forward pass, with their appearance attributes obtained through decoupled appearance modeling. (b) Leveraging the strong observation provided by the input image itself, the zeroth-order SH coefficient of each pixel-aligned Gaussian is directly derived from its corresponding pixel RGB value. (c) Higher-order SH coefficients are predicted by a view-dependent appearance head, which takes two effective 3D-aware inductive biases, GVSRE and WCM, enabling more accurate modeling of view-dependent appearance.}
    \label{fig:pipeline}
    \phantomsubcaption\label{fig:pipeline_a}
    \phantomsubcaption\label{fig:pipeline_b}
    \phantomsubcaption\label{fig:pipeline_c}
    % \label{fig:pipeline}
\end{figure}

We propose \textbf{AESplat}, a novel and general pose-free feed-forward 3DGS framework that employs a decoupled strategy to predict view-independent and view-dependent Gaussian appearances separately. Given a pair of unposed images, our method jointly predicts Gaussian primitives in a canonical space together with the corresponding camera parameters. For view-independent appearance, we leverage a training-free strategy that directly drives the zeroth-order SH coefficient from the input images. Based on the predicted relative camera poses, we develop two effective 3D-aware inductive biases, GVSRE and WCM, to guide view-dependent appearance modeling through a shallow MLP. Through this effective decoupled appearance modeling strategy, our method produces a higher-quality Gaussian representation, resulting in better novel-view rendering.

The overall pipeline is illustrated in Fig.~\ref{fig:pipeline}. We first present the motivation for our decoupled appearance modeling strategy in Sec.~\ref{method:motivation}. We then describe in detail the proposed strategy in Sec.~\ref{method:vid} and Sec.~\ref{method:vd}. Finally, the overall training objectives are described in Sec.~\ref{method:loss}.

\subsection{Motivation}
\label{method:motivation}

Our motivation stems from a detailed analysis of SH. SH forms an orthogonal basis for functions defined on the sphere. 
In vanilla 3DGS~\citep{kerbl20233d}, the appearance of each Gaussian $\mathcal{G}$ is represented by a linear combination of SH basis functions:
\begin{equation}
\label{eq:sh}
{\displaystyle \vC}_{\mathcal{G}}({\displaystyle \vd}) =
\operatorname{max}\Big(
\underbrace{0.5 + {\displaystyle \vc}_0^0 Y_0^0}
_{\text{view-independent appearance}}
+
\underbrace{
\textstyle\sum_{l=1}^{L}\sum_{m=-l}^{l}
{\displaystyle \vc}_l^m Y_l^m({\displaystyle \vd})
}_{\text{view-dependent appearance}}, \quad 0
\Big).
\end{equation}
where $\displaystyle \vd$ denotes the viewing direction, $Y_l^m(\cdot)$ denotes the real SH basis function of degree $l$ and order $m$, and $\displaystyle \vc_l^m \in {\displaystyle \R}^3$ is the corresponding learnable SH coefficient. In particular, the zeroth-order SH basis is a constant given by $Y_0^0 = \frac{1}{2\sqrt{\pi}}$, which is independent of the viewing direction. In contrast, higher-order SH basis functions are dependent on the viewing direction. This indicates that the zeroth-order coefficient corresponds to view-independent appearance, while higher-order SH terms correspond to view-dependent effects. Motivated by this distinction, we argue that predicting SH coefficients corresponding to different appearances in the same manner is suboptimal. Therefore, we adopt a decoupled approach that models the two components with tailored strategies.

\subsection{View-independent Appearance Modeling (VIAM)}
\label{method:vid}
For the basic view-independent appearance represented by the zeroth-order SH coefficient, we observe that the pixel corresponding to each Gaussian provides strong direct information. Rather than designing an additional network to predict it, we directly derive the zeroth-order SH coefficient of each pixel-aligned Gaussian from its corresponding pixel RGB value. Specifically, given a pixel color ${\displaystyle \vC}^{v}_j\in {\displaystyle \R}^3$ for pixel $j$ in context view $v$, we convert it to the zeroth-order SH coefficient by following the initialization strategy of the vanilla 3DGS~\citep{kerbl20233d}:  
\begin{equation}
    {\displaystyle \vc}^{0}_0(v,j)
    =
    \frac{{\displaystyle \vC}^{v}_j-0.5}{Y^0_0}.
\end{equation}
This strategy, which we term Image-to-Direct Component (I2DC),  provides a reliable estimation for the zeroth-order appearance component while avoiding the need to relearn color information already present in the input images. As shown in the ablation study in Sec.~\ref{ablation}, the model is still able to achieve comparable rendering quality without higher-order SH coefficients. 

\subsection{View-dependent Appearance Modeling (VDAM)}
\label{method:vd}

As discussed above, higher-order SH coefficients are closely tied to view-dependent appearance. We therefore argue that explicit view information is essential for accurately estimating these coefficients. To better capture such complex view-dependent effects, we introduce a view-dependent appearance prediction head to estimate the higher-order SH coefficients. Unlike previous approaches that directly encode the ground-truth camera poses or convert them into Plücker ray embeddings for interaction within a Transformer, we instead design two effective 3D-aware inductive biases based on the predicted camera parameters and provide them as inputs to the appearance prediction head. These biases encode the 3D spatial relationships between Gaussians and different viewpoints, as well as cross-view color information. This enables the network to more effectively model appearance variations under different viewing directions.

% As discussed above, higher-order SH coefficients characterize view-dependent appearance, suggesting that explicit view information is crucial for their accurate estimation. To better capture such complex view-dependent effects, we employ a view-dependent appearance head to estimate the higher-order SH coefficients. More importantly, we introduce two novel 3D-aware inductive biases, derived from the predicted camera parameters, and use them as inputs to the appearance head. These biases explicitly encode the underlying 3D geometric and view-dependent relationships, enabling the network to better model appearance variations across different viewing directions.

\noindent\textbf{Gaussian-to-Views Spatial Relation Embedding (GVSRE)}. As illustrated in Fig.~\ref{fig:pipeline_c}, we model the observation of each Gaussian from different viewpoints through the spatial relationship between its center and the corresponding viewpoint. For each Gaussian and viewpoint center, we adopt a compact representation~\citep{jeong2026viewsplat}, consisting of a 3D directional vector and a scalar log-distance value. Specifically, for each Gaussian center ${\displaystyle \vmu}^{v_s}_{j}$ from view $v_s\!\in\!\mathcal{I}_\mathcal{C}$, we form a multi-view spatial descriptor by concatenating its spatial relationship with each context camera center ${\displaystyle \vo}^{v_c}\in {\displaystyle \R}^3$:
\begin{equation}
\mathbf{r}^{v_s}_j
=
\big\|_{v_c\in\mathcal{I}_\mathcal{C}}
\!\left[
\frac{{\displaystyle \vo}^{v_c}\!-\!{\displaystyle \vmu}^{v_s}_{j}}{\|{\displaystyle \vo}^{v_c}\!-\!{\displaystyle \vmu}^{v_s}_{j}\|_2},\;
\log\|{\displaystyle \vo}^{v_c}\!-\!{\displaystyle \vmu}^{v_s}_{j}\|_2
\right]
\in {\displaystyle \R}^{4|\mathcal{V}_\mathcal{C}|},
\end{equation}
where $\big\|$ denotes concatenation over all context views. A shallow MLP then maps each $\mathbf{r}^{v_s}_j$ to a spatial relation embedding, GVSRE. Unlike the fine-tune method~\citep{jeong2026viewsplat}, which encodes the pose feature only with respect to the target view, our GVSRE aggregates spatial relationships from all context cameras, providing a globally consistent geometric cue for appearance prediction.

\noindent\textbf{Warped Color Map (WCM)}. In addition to GVSRE, we introduce a novel module that explicitly provides the view-dependent appearance head with the observed color of each Gaussian from the other context view, providing complementary appearance cue for view-dependent appearance prediction. Concretely, we project each source Gaussian center into the neighboring context view and sample its RGB observation. For a context view ${v_p} \in {\mathcal{I}_\mathcal{C}}$, the projected coordinate of Gaussian $\mathcal{G}_j$ in other view ${v_q} \in {\mathcal{I}_\mathcal{C}}$ is
\begin{equation}
    {\displaystyle \vu}_{j}^{{v_q}\leftarrow{v_p}}
    =
    \pi\left(
    {\displaystyle \mK}^{v_q}
    {\displaystyle \mT}^{{v_q}\leftarrow{v_p}}
    {\displaystyle \vmu}^{v_p}_j
    \right),
\end{equation}
where ${\displaystyle \mK}^{v_q}$ denotes the intrinsic matrix of view $v_q$, 
${\displaystyle \mT}^{{v_q}\leftarrow{v_p}}$ denotes the camera-to-camera transformation from view $v_p$ to $v_q$, 
and $\pi(\cdot)$ denotes the perspective projection. The warped color $\hat{{\displaystyle \vC}}_{j}^{{v_q}\leftarrow{v_p}}$ is obtained from ${\displaystyle \vI}^{v_q}$ via bilinear interpolation at ${\displaystyle \vu}_{j}^{{v_q}\leftarrow{v_p}}$.
We also compute a binary validity mask:
\begin{equation}
    m_{j}^{{v_q}\leftarrow{v_p}}
    =
    \mathbbm{1}
    \left[
    z_{j}^{{v_q}\leftarrow{v_p}} > 0
    \right]
    \cdot
    \mathbbm{1}
    \left[
    u_{j}^{{v_q}\leftarrow{v_p}} \in \Omega
    \right],
\end{equation}
where $z_{j}^{{v_q}\leftarrow{v_p}}$ denotes the depth of the projected point in view $v_q$, and $\Omega$ denotes the normalized image domain. Consequently, $m_{j}^{{v_q}\leftarrow{v_p}}=1$ only if the projected point lies in front of the camera and falls within the image boundaries. Aggregating the warped colors and validity masks of all $H\times W$ Gaussians in view $v_p$ produces the warped color map $\hat{{\displaystyle \vC}}^{{v_q}\leftarrow{v_p}}$ and the corresponding validity mask ${\displaystyle \vM}^{{v_q}\leftarrow{v_p}}$ in view $v_q$, which together form the WCM.

\noindent\textbf{View-dependent Appearance Prediction.}
We employ a four-layer MLP as the view-dependent appearance head, with GELU~\citep{hendrycks2016gaussian} activations after the first three hidden layers and a linear output layer to predict view-dependent appearance. It takes the GVSRE, the WCM, and the intermediate features from the Gaussian geometry head as input. After layer normalization, the view-dependent appearance head predicts the higher-order SH coefficients. 

\subsection{Loss Function}
\label{method:loss}
During training, the model is optimized solely using the
photometric error between the rendered image and the corresponding
ground-truth image at the target view. The training objective combines the MSE and LPIPS~\citep{zhang2018unreasonable} losses, formulated as
\begin{equation}
\mathcal{L}
=
\lambda_{\mathrm{mse}}
\left\|{\displaystyle \vI}_{\mathrm{render}}-{\displaystyle \vI}_{\mathrm{gt}}\right\|_2^2
+
\lambda_{\mathrm{lpips}}
\operatorname{LPIPS}
\!\left({\displaystyle \vI}_{\mathrm{render}}, {\displaystyle \vI}_{\mathrm{gt}}\right),
\label{eq:loss}
\end{equation}
where ${\displaystyle \vI}_{\mathrm{render}}$ and ${\displaystyle \vI}_{\mathrm{gt}}$ denote the rendered
image and the corresponding ground-truth image, respectively, and
$\lambda_{\mathrm{mse}}$ and $\lambda_{\mathrm{lpips}}$ are the
corresponding loss weights.

\section{Experiments}
\label{experiment}

\subsection{Experimental Settings}

\noindent\textbf{Datasets.} We train and evaluate our method on RealEstate10K~\citep{10.1145/3197517.3201323} and ACID~\citep{liu2021infinite} with the official train/test splits used in prior work~\citep{ye2025no,huang2025no}. RealEstate10K consists of large-scale indoor real-estate tour videos, whereas ACID covers outdoor natural environments filmed from aerial platforms. To evaluate cross-dataset generalization, we additionally evaluate on ACID, the large-scale outdoor dataset DL3DV~\citep{ling2024dl3dv}, and the indoor dataset ScanNet++~\citep{yeshwanth2023scannet++}.

\noindent\textbf{Baselines.} We compare our method against several SOTA approaches, including pose-required methods (pixelSplat~\citep{charatan2024pixelsplat}, MVSplat~\citep{chen2024mvsplat}, DepthSplat~\citep{xu2025depthsplat}, and YoNoSplat~\citep{ye2026yonosplat}), supervised pose-free methods (Splatt3R~\citep{smart2024splatt3r} and NoPoSplat~\citep{ye2025no}), and self-supervised pose-free methods (SelfSplat~\citep{kang2025selfsplat}, PF3Splat~\citep{hong2025pfplat}, SPFSplat~\citep{huang2025no}, SPFSplatV2-L~\citep{huang2025spfsplatv2}, and NAS3R~\citep{huang2026none}). 

% All metrics reported for NoPoSplat are obtained with evaluation-time pose alignment~\citep{ye2025no}.

% Pose-required methods render target views using ground-truth poses, while some pose-free methods use predicted target poses (e.g., PF3Splat, SelfSplat, SPFSplat, and SPFSplatV2) or adopt evaluation-time pose alignment (EPA) (e.g., NoPoSplat), where the reconstructed Gaussians are fixed and only the target-view pose is optimized at test time. For a fair evaluation, our method renders all target views using the predicted camera poses. 在补充材料里说
\noindent\textbf{Evaluation Protocol.} To evaluate performance, we report commonly used metrics for novel-view rendering quality, including PSNR (dB), SSIM~\citep{wang2004image}, and LPIPS~\citep{zhang2018unreasonable}. 

\noindent\textbf{Implementation Details}. AESplat is implemented in PyTorch, and all models are trained on a single NVIDIA A6000 GPU. Following NAS3R~\citep{huang2026none}, we adopt the same training setup, where each sample consists of two context views and one target view at an input resolution of $224\times224$. We use the AdamW~\citep{loshchilov2019iclr-decoupled} optimizer with a learning rate of $3\times10^{-4}$ for the view-dependent appearance head, while keeping the same learning rate as NAS3R for the remaining modules. We use a batch size of 10 and train for 400k iterations. $\lambda_{\mathrm{mse}}=1$, $\lambda_{\mathrm{lpips}}=0.05$. Additional implementation details are provided in the supplementary material.

\subsection{Results}
\begin{table}[t]
\centering
\caption{\textbf{Performance comparison of novel view synthesis on RealEstate10K and ACID
datasets}. We report the average metrics across all test scenes. The \textbf{best} and \underline{second-best} results are highlighted. $-$ indicates that the result was not reported in the original paper.}

\fontsize{8}{9}\selectfont
\setlength{\tabcolsep}{6pt}        % 列间距（默认约6pt）

\begin{tabular}{l ccc ccc}
\toprule
\multirow{2}{*}{\textbf{Method}}
& \multicolumn{3}{c}{\textbf{RealEstate10K}}
& \multicolumn{3}{c}{\textbf{ACID}} \\
\cmidrule(lr){2-4}
\cmidrule(lr){5-7}
& PSNR $\uparrow$ & SSIM $\uparrow$ & LPIPS $\downarrow$
& PSNR $\uparrow$ & SSIM $\uparrow$ & LPIPS $\downarrow$ \\
\midrule

\rowcolor{gray!20}
\multicolumn{7}{l}{\textit{Supervised Pose-required}} \\
pixelSplat~\citep{charatan2024pixelsplat}
& 23.859 & 0.808 & 0.184
& 25.889 & 0.780 & 0.194 \\
MVSplat~\citep{chen2024mvsplat}
& 24.012 & 0.812 & 0.175
& 25.561 & 0.775 & 0.195 \\
DepthSplat~\citep{xu2025depthsplat}
& 25.595 & 0.852 & 0.145 & $-$ & $-$ & $-$ \\
YoNoSplat~\citep{ye2026yonosplat}
& 24.233 & 0.813 & 0.162 & $-$ & $-$ & $-$ \\
% & 25.383 & 0.773 & 0.195 \\

\rowcolor{gray!20}
\multicolumn{7}{l}{\textit{Supervised Pose-free}} \\
Splatt3R~\citep{smart2024splatt3r}
& 18.688 & 0.337 & 0.596 & 18.060 & 0.510 & 0.407 \\
NoPoSplat ~\citep{ye2025no}
& 25.033 & 0.838 & 0.160
& 25.961 & 0.781 & 0.189 \\

\rowcolor{gray!20}
\multicolumn{7}{l}{\textit{Self-Supervised Pose-free}} \\
SelfSplat~\citep{kang2025selfsplat}
& 19.152 & 0.680 & 0.328 & 22.089 & 0.694 & 0.298 \\
PF3Splat~\citep{hong2025pfplat} & 21.042 & 0.739 & 0.233 & 21.206 & 0.632 & 0.293 \\
SPFSplat~\citep{huang2025no}
& 25.484 & 0.847 & 0.153 & 26.070 & 0.781 & 0.186\\
SPFSplatV2-L~\citep{huang2025spfsplatv2}
& 25.668 & 0.855 & 0.137 & 26.674 & 0.806 & 0.162 \\
NAS3R~\citep{huang2026none}
& \underline{25.888} & \underline{0.861} & \underline{0.136} & \underline{26.832} & \underline{0.813} & \underline{0.160} \\
\textbf{AESplat} (Ours)
& \textbf{26.691} & \textbf{0.872} & \textbf{0.128} & \textbf{27.679} & \textbf{0.825} & \textbf{0.151} \\

\bottomrule
\end{tabular}
\label{tab:main_results}
\end{table}
\begin{figure}[t!]
    \centering 
    \includegraphics[width=\textwidth]{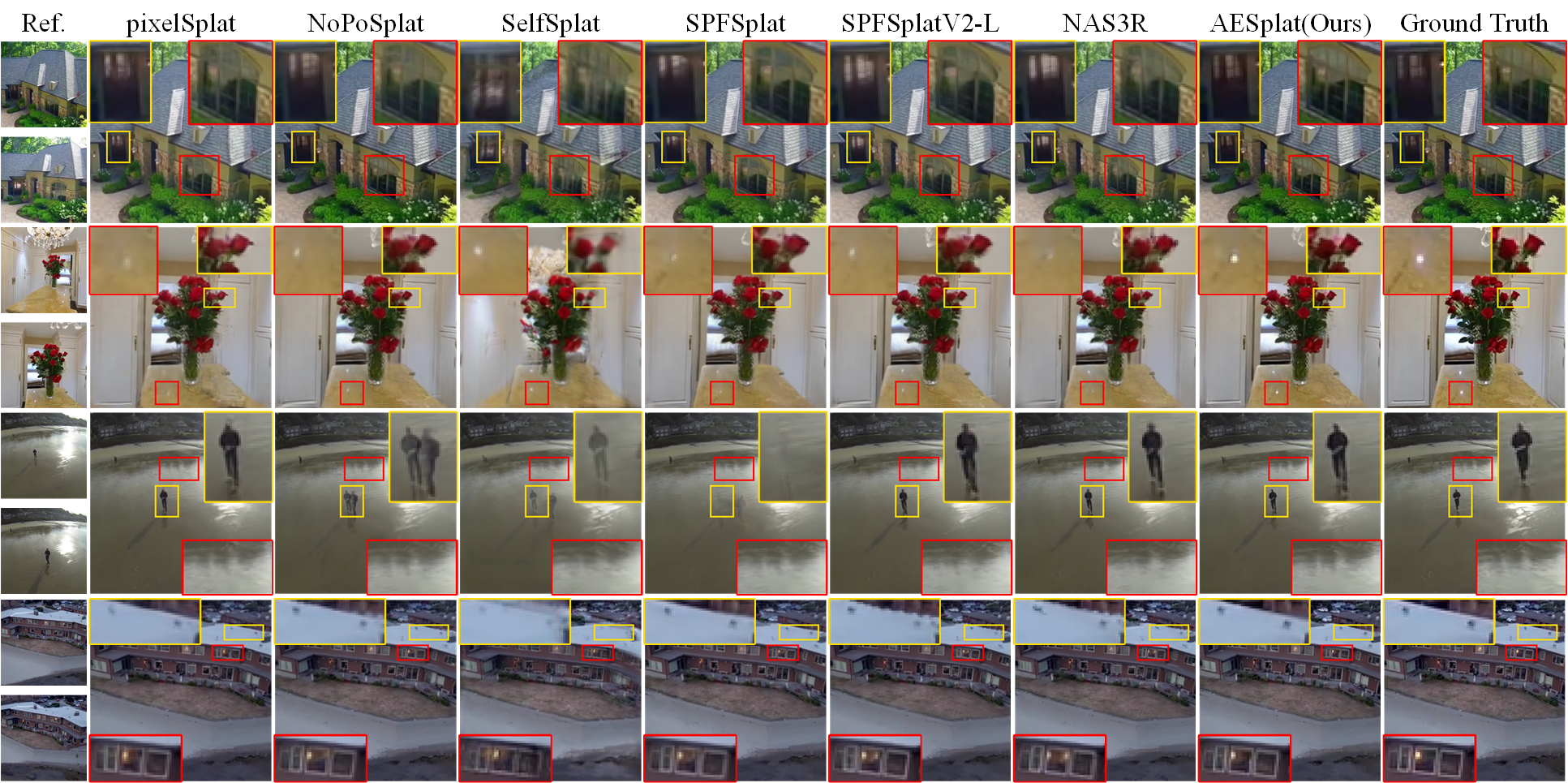} 
    \caption{\textbf{Qualitative results of Tab.~\ref{tab:main_results}}. The leftmost column shows the two-view context images. The top two rows are from RealEstate10K, and the bottom two are from ACID.}
    \label{fig:performance}
    % \vspace{-2mm}
\end{figure}
\noindent\textbf{Novel View Synthesis}. As shown in Tab.~\ref{tab:main_results}, AESplat achieves new SOTA performance across all evaluation metrics on both RealEstate10K and ACID. Compared with NAS3R, our method improves PSNR by 0.803~dB on RealEstate10K and 0.847~dB on ACID, with consistent improvements in SSIM and LPIPS. The qualitative comparisons in Fig.~\ref{fig:performance} show that our method consistently recovers finer, sharper, and more faithful appearance details across diverse scenes, demonstrating the effectiveness of our decoupled appearance modeling strategy.

\begin{figure}[t]
    \centering 
    \includegraphics[width=\textwidth]{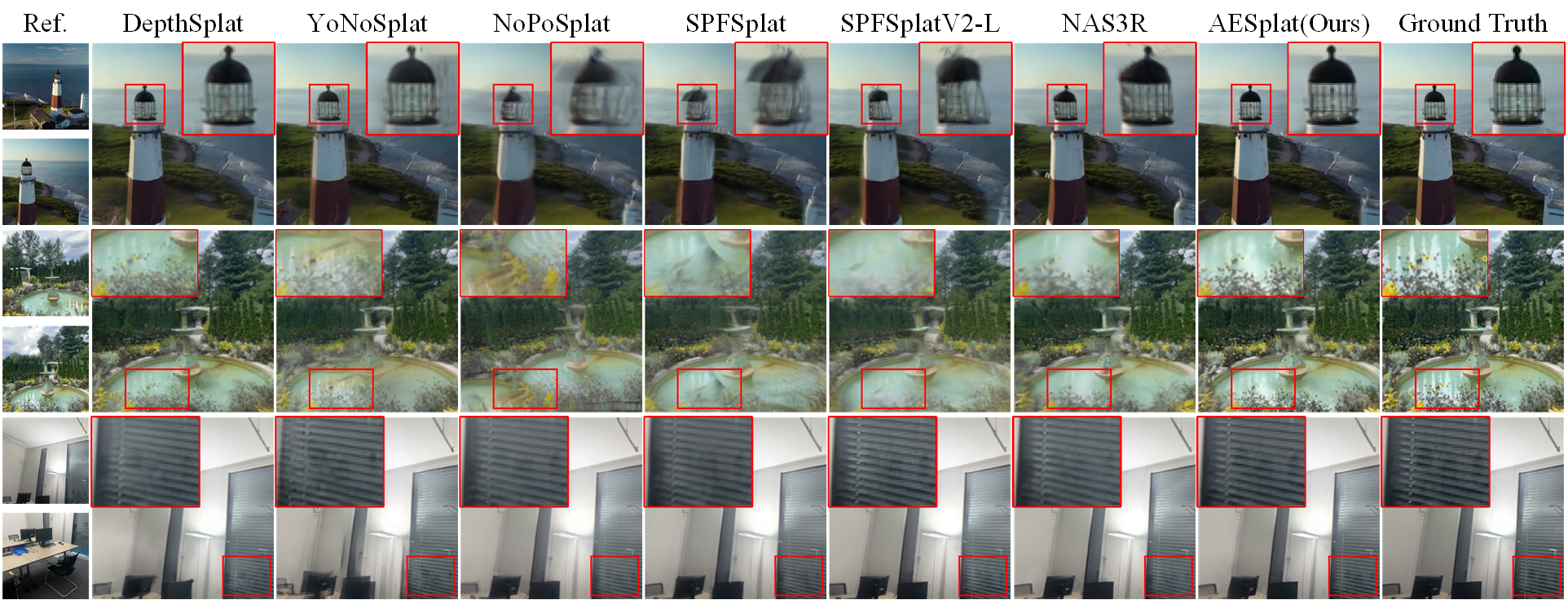} 
    \caption{\textbf{Qualitative results of Tab.~\ref{tab:cross_dataset_results}.} From top to bottom, the three rows show zero-shot results on ACID, DL3DV, and ScanNet++ when trained on RealEstate10K.}
    \label{fig:cross_dataset}
\end{figure}

\noindent\textbf{Cross-Dataset Generalization}. 
To evaluate our method under cross-dataset distribution shifts, we train exclusively on the RealEstate10K dataset and perform zero-shot evaluation on ACID, DL3DV, and ScanNet++. As shown in Tab.~\ref{tab:cross_dataset_results} and Fig.~\ref{fig:cross_dataset}, our method achieves the best overall performance across all three datasets, outperforming existing methods on nearly all metrics. These results demonstrate that our proposed strategy learns transferable representations that generalize beyond the training distribution of RealEstate10K. 

\begin{table}[t]
\centering
\caption{\textbf{Cross-dataset generalization}. All methods are trained on RealEstate10K and
evaluated in a zero-shot setting on ACID, DL3DV, and ScanNet++. }

\fontsize{8}{9}\selectfont
\setlength{\tabcolsep}{5pt}

\begin{tabular}{l ccc ccc ccc}
    \toprule
\multirow{2}{*}{\textbf{Method}}
& \multicolumn{3}{c}{\textbf{ACID}}
& \multicolumn{3}{c}{\textbf{DL3DV}}
& \multicolumn{3}{c}{\textbf{ScanNet++}}\\
\cmidrule(lr){2-4}
\cmidrule(lr){5-7}
\cmidrule(lr){8-10}
& PSNR $\uparrow$ & SSIM $\uparrow$ & LPIPS $\downarrow$
& PSNR $\uparrow$ & SSIM $\uparrow$ & LPIPS $\downarrow$
& PSNR $\uparrow$ & SSIM $\uparrow$ & LPIPS $\downarrow$ \\
\midrule
    \rowcolor{gray!20} \multicolumn{10}{l}{\textit{Supervised Pose-required}} \\ 
pixelSplat  & 25.477  & 0.770 & 0.207 & 18.688 & 0.582 & 0.354 & 18.422 & 0.720 & 0.278 \\
MVSplat & 25.525  & 0.773 & 0.199 & 17.786 & 0.545 & 0.357 & 17.138 & 0.687 & 0.297  \\
DepthSplat & 26.012 & 0.791 & 0.185 & 19.553 & 0.611 & 0.285 & 20.775 & 0.760 & 0.254 \\
YoNoSplat & 24.246 & 0.721 & 0.222 & 19.636 & 0.594 & 0.311 & 21.075 & 0.744 & 0.254 \\
\rowcolor{gray!20} \multicolumn{10}{l}{\textit{Supervised Pose-free}} \\ 
NoPoSplat & 25.764  & 0.776 & 0.199 & 19.974 & 0.612 & 0.305 & \textbf{22.136} & 0.798 & 0.232  \\
    \rowcolor{gray!20} \multicolumn{10}{l}{\textit{Self-Supervised Pose-free}} \\ 
    SelfSplat &  22.204 & 0.686 & 0.316 & 15.047 & 0.410 & 0.498 & 13.277 & 0.538 & 0.534 \\
    SPFSplat &  25.965 & 0.781 & 0.190 & 19.172 & 0.573 & 0.315 & 19.971 & 0.738 & 0.265 \\
    SPFSplatV2-L &  26.361 & 0.796 & 0.169 & 19.743 & 0.613 & 0.277 & 21.796 & \underline{0.811} & \underline{0.200} \\
    NAS3R & \underline{26.663}  & \underline{0.807} & \underline{0.166}  & \underline{19.842} & \underline{0.628} & \underline{0.274} & 21.028 & 0.799 & 0.210 \\
    \textbf{AESplat} (Ours) & \textbf{27.375}  & \textbf{0.819} & \textbf{0.156} & \textbf{20.430} & \textbf{0.648} & \textbf{0.259} & \underline{21.892} & \textbf{0.812} & \textbf{0.199}   \\
    \bottomrule
    \end{tabular}
\label{tab:cross_dataset_results}
\end{table}

\begin{figure}[t!]
    \centering

    % Left: Table
    \begin{minipage}[c]{0.58\textwidth}
    \centering
    \captionof{table}{\textbf{Baseline Generality.} Our approach consistently improves the performance of different Baselines.}
    \label{tab:backbone_generality}
    
    {\fontsize{8pt}{8pt}\selectfont
    \setlength{\tabcolsep}{10pt}
    \begin{tabular}{lccc}
        \toprule
        \multirow{2}{*}{\textbf{Method}} & 
        \multicolumn{3}{c}{\textbf{RealEstate10K}} \\
        \cmidrule(lr){2-4}
        & PSNR $\uparrow$ & SSIM $\uparrow$ & LPIPS $\downarrow$ \\
        \midrule
        NAS3R-m & 25.814 & 0.856 & 0.149 \\
        +Ours & \textbf{26.294} & \textbf{0.860} & \textbf{0.141} \\
        \midrule
        SPFSplatV2 & 25.693 & 0.853 & 0.149 \\
        +Ours & \textbf{26.181} & \textbf{0.859} & \textbf{0.141} \\
        \bottomrule
    \end{tabular}
    }
\end{minipage}
    \hfill
    % Right: Figure
    \begin{minipage}[c]{0.40\textwidth}
        \centering
        \includegraphics[
            width=\textwidth,
            height=3.0cm,
            keepaspectratio
        ]{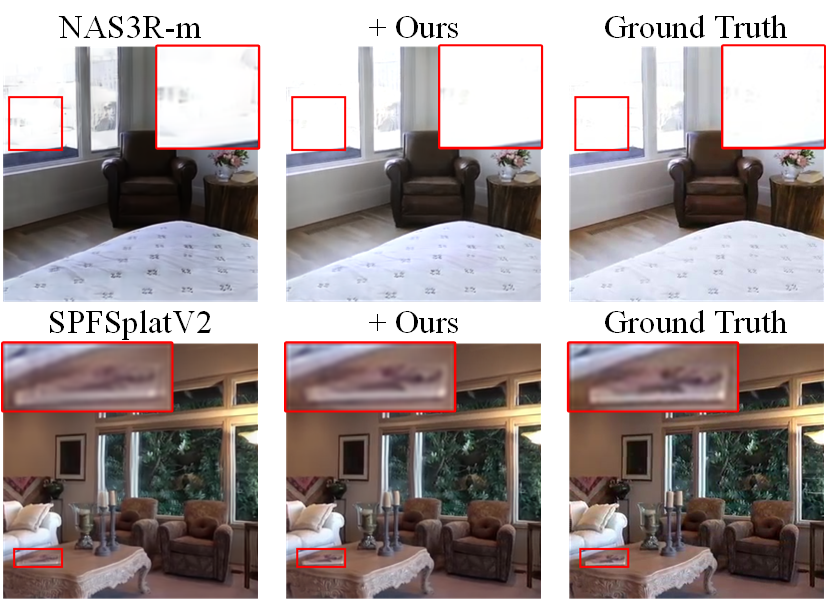}
        \captionof{figure}{\textbf{Qualitative results of Tab.~\ref{tab:backbone_generality}.}}
        \label{fig:backbone_generality}
    \end{minipage}

\end{figure}

\noindent\textbf{Baseline Generality.} To evaluate the generality of our approach, we further apply our decoupled appearance modeling strategy to NAS3R-m and SPFSplatV2, both of which use MASt3R as their backbone. As shown in Tab.~\ref{tab:backbone_generality} and Fig.~\ref{fig:backbone_generality}, our strategy consistently improves rendering quality across different baselines, demonstrating its strong adaptability and generalizability to diverse methods.

\subsection{Ablation Studies}
\label{ablation}

\begin{figure}[t]
    \centering 
    \includegraphics[width=\textwidth]{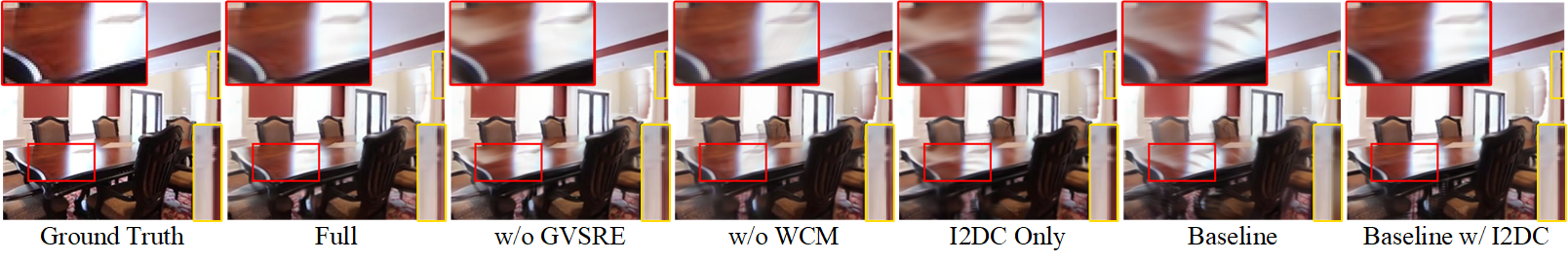} 
    \caption{\textbf{Qualitative results of Tab.~\ref{tab:ablation}.} GVSRE and WCM model view-dependent appearance variations (red boxes), while I2DC better preserves the base appearance component (yellow boxes).}
    \label{fig:ablation}
\end{figure}

We provide results on ablation studies in Tab.~\ref{tab:ablation} and Fig.~\ref{fig:ablation}. For a fair comparison, all models are trained on RealEstate10K for 50K steps, following SelfSplat~\citep{kang2025selfsplat}.

\noindent\textbf{No GVSRE}.
GVSRE encodes the spatial relationships between each Gaussian and all context views, providing the appearance head with explicit cues about how the Gaussian is observed from different viewpoints. As shown in Tab.~\ref{tab:ablation}, removing GVSRE causes a $0.269$~dB drop in PSNR. Qualitatively, the reflection on the tabletop becomes noticeably blurred, indicating that the spatial cues provided by GVSRE are important for faithfully modeling view-dependent details.

\noindent\textbf{No WCM}.
WCM provides cross-view appearance observations of each Gaussian by warping its color to other context views, offering complementary evidence for capturing view-dependent variations. As shown in Tab.~\ref{tab:ablation} and Fig.~\ref{fig:ablation}, removing WCM causes a $0.256$~dB drop in PSNR, while the tabletop reflection is visibly blurred in the qualitative comparison.

\noindent\textbf{I2DC}.
We remove the view-dependent branch and retain only the view-independent appearance modeled by I2DC. As shown in Tab.~\ref{tab:ablation}, this variant achieves comparable rendering quality to the baseline, with only a $0.144$~dB gap in PSNR, validating the effectiveness of I2DC. Replacing the zeroth-order SH coefficients in the baseline with those derived by I2DC further improves PSNR by $0.154$~dB and yields more faithful reconstruction of wall appearance. These results demonstrate that I2DC effectively models view-independent appearance.

\subsection{Limitations}

Despite the strong rendering quality achieved by our method, several limitations remain. 
\begin{wraptable}{r}{0.5\textwidth}
    \centering
    \caption{\textbf{Ablations.} We evaluate the contribution of the proposed method on RealEstate10K.}
    \label{tab:ablation}
    \begin{scriptsize}
        \resizebox{\linewidth}{!}{
            \begin{tabular}{l | ccc}
                \toprule
                \textbf{Method} & PSNR$\uparrow$ & SSIM$\uparrow$ & LPIPS$\downarrow$ \\ 
                \midrule
                Full & \textbf{24.268} & \textbf{0.821} & \textbf{0.160} \\ 
                \midrule
                (1) w/o GVSRE & 23.999 & 0.816 & 0.163 \\
                (2) w/o WCM & 24.012 & 0.818 & 0.161 \\
                (3) I2DC only & 23.610 & 0.808 & 0.173 \\
                (4) Baseline (NAS3R) & 23.754 & 0.808 & 0.171 \\
                (5) Baseline (NAS3R) + I2DC & 23.908 & 0.813 & 0.165 \\ 
                \bottomrule
            \end{tabular}
        }
    \end{scriptsize}
\end{wraptable}
First, our method inherits an inherent limitation of the feed-forward NVS paradigm: content that is completely absent from the context views cannot be reliably reconstructed, as no visual evidence is available for such regions. Second, I2DC directly derives the DC component from the observed RGB values, which only approximates view-independent appearance under severe exposure variations and non-Lambertian specularities. Future work could explore more accurate estimation and disentanglement of view-independent appearance.

 % Future work could explore incorporating ground-truth or more accurate poses during training or fine-tuning to further improve reconstruction and rendering quality.

\section{Conclusion}
In summary, we present a novel and general framework for pose-free feed-forward 3DGS that shifts from unified to decoupled appearance modeling, substantially improving rendering quality. We introduce a decoupled appearance modeling strategy that directly obtains zeroth-order SH coefficient from input images without training, while leveraging two effective 3D-aware inductive biases to better predict higher-order SH coefficients. Extensive experiments demonstrate the effectiveness and generality of our approach, which consistently achieves state-of-the-art rendering quality among existing pose-free methods. We believe our work provides a new perspective on appearance modeling and offers an effective solution for high-fidelity pose-free feed-forward 3D Gaussian reconstruction.

\bibliography{iclr2026_conference}

@article{kerbl20233d,
  title={3d gaussian splatting for real-time radiance field rendering.},
  author={Kerbl, Bernhard and Kopanas, Georgios and Leimk{\"u}hler, Thomas and Drettakis, George and others},
  journal={ACM Trans. Graph.},
  volume={42},
  number={4},
  pages={139--1},
  year={2023}
}

@inproceedings{wang2025vggt,
  title={Vggt: Visual geometry grounded transformer},
  author={Wang, Jianyuan and Chen, Minghao and Karaev, Nikita and Vedaldi, Andrea and Rupprecht, Christian and Novotny, David},
  booktitle={Proceedings of the Computer Vision and Pattern Recognition Conference},
  pages={5294--5306},
  year={2025}
}

@inproceedings{leroy2024grounding,
  title={Grounding image matching in 3d with mast3r},
  author={Leroy, Vincent and Cabon, Yohann and Revaud, J{\'e}r{\^o}me},
  booktitle={European conference on computer vision},
  pages={71--91},
  year={2024},
  organization={Springer}
}

@inproceedings{charatan2024pixelsplat,
  title={pixelsplat: 3d gaussian splats from image pairs for scalable generalizable 3d reconstruction},
  author={Charatan, David and Li, Sizhe Lester and Tagliasacchi, Andrea and Sitzmann, Vincent},
  booktitle={Proceedings of the IEEE/CVF conference on computer vision and pattern recognition},
  pages={19457--19467},
  year={2024}
}

@inproceedings{chen2024mvsplat,
  title={Mvsplat: Efficient 3d gaussian splatting from sparse multi-view images},
  author={Chen, Yuedong and Xu, Haofei and Zheng, Chuanxia and Zhuang, Bohan and Pollefeys, Marc and Geiger, Andreas and Cham, Tat-Jen and Cai, Jianfei},
  booktitle={European conference on computer vision},
  pages={370--386},
  year={2024},
  organization={Springer}
}

@inproceedings{ye2025no,
  title={No pose, no problem: Surprisingly simple 3d gaussian splats from sparse unposed images},
  author={Ye, Botao and Liu, Sifei and Xu, Haofei and Li, Xueting and Pollefeys, Marc and Yang, Ming-Hsuan and Peng, Songyou},
  booktitle={International Conference on Learning Representations},
  volume={2025},
  pages={54009--54033},
  year={2025}
}

@article{jiang2025anysplat,
  title={Anysplat: Feed-forward 3d gaussian splatting from unconstrained views},
  author={Jiang, Lihan and Mao, Yucheng and Xu, Linning and Lu, Tao and Ren, Kerui and Jin, Yichen and Xu, Xudong and Yu, Mulin and Pang, Jiangmiao and Zhao, Feng and others},
  journal={ACM Transactions on Graphics (TOG)},
  volume={44},
  number={6},
  pages={1--16},
  year={2025},
  publisher={ACM New York, NY, USA}
}

@inproceedings{huang2026none,
  title={From None to All: Self-Supervised 3D Reconstruction via Novel View Synthesis},
  author={Huang, Ranran and Luo, Weixun and Mao, Ye and Mikolajczyk, Krystian},
  booktitle={Proceedings of the IEEE/CVF Conference on Computer Vision and Pattern Recognition},
  pages={37358--37369},
  year={2026}
}

@inproceedings{huang2025no,
  title={No pose at all: Self-supervised pose-free 3d gaussian splatting from sparse views},
  author={Huang, Ranran and Mikolajczyk, Krystian},
  booktitle={Proceedings of the IEEE/CVF International Conference on Computer Vision},
  pages={27947--27957},
  year={2025}
}

@inproceedings{zhao2026rayzer,
  title={E-rayzer: Self-supervised 3d reconstruction as spatial visual pre-training},
  author={Zhao, Qitao and Tan, Hao and Wang, Qianqian and Bi, Sai and Zhang, Kai and Sunkavalli, Kalyan and Tulsiani, Shubham and Jiang, Hanwen},
  booktitle={Proceedings of the IEEE/CVF Conference on Computer Vision and Pattern Recognition},
  pages={7525--7535},
  year={2026}
}

@inproceedings{kang2025selfsplat,
  title={Selfsplat: Pose-free and 3d prior-free generalizable 3d gaussian splatting},
  author={Kang, Gyeongjin and Yoo, Jisang and Park, Jihyeon and Nam, Seungtae and Im, Hyeonsoo and Shin, Sangheon and Kim, Sangpil and Park, Eunbyung},
  booktitle={Proceedings of the Computer Vision and Pattern Recognition Conference},
  pages={22012--22022},
  year={2025}
}

@inproceedings{
hong2025pfplat,
title={{PF}3plat: Pose-Free Feed-Forward 3D Gaussian Splatting for Novel View Synthesis},
author={Sunghwan Hong and Jaewoo Jung and Heeseong Shin and Jisang Han and Jiaolong Yang and Chong Luo and Seungryong Kim},
booktitle={Forty-second International Conference on Machine Learning},
year={2025}
}

@inproceedings{xu2025depthsplat,
  title={Depthsplat: Connecting gaussian splatting and depth},
  author={Xu, Haofei and Peng, Songyou and Wang, Fangjinhua and Blum, Hermann and Barath, Daniel and Geiger, Andreas and Pollefeys, Marc},
  booktitle={Proceedings of the Computer Vision and Pattern Recognition Conference},
  pages={16453--16463},
  year={2025}
}

@article{huang2025spfsplatv2,
  title={SPFSplatV2: Efficient Self-Supervised Pose-Free 3D Gaussian Splatting from Sparse Views},
  author={Huang, Ranran and Mikolajczyk, Krystian},
  journal={arXiv preprint arXiv:2509.17246},
  year={2025}
}

@article{10.1145/3197517.3201323,
author = {Zhou, Tinghui and Tucker, Richard and Flynn, John and Fyffe, Graham and Snavely, Noah},
title = {Stereo magnification: learning view synthesis using multiplane images},
year = {2018},
issue_date = {August 2018},
publisher = {Association for Computing Machinery},
address = {New York, NY, USA},
volume = {37},
number = {4},
issn = {0730-0301},
journal = {ACM Trans. Graph.},
month = jul,
articleno = {65},
numpages = {12}
}

@inproceedings{dosovitskiyimage,
  title={An Image is Worth 16x16 Words: Transformers for Image Recognition at Scale},
  author={Dosovitskiy, Alexey and Beyer, Lucas and Kolesnikov, Alexander and Weissenborn, Dirk and Zhai, Xiaohua and Unterthiner, Thomas and Dehghani, Mostafa and Minderer, Matthias and Heigold, Georg and Gelly, Sylvain and others},
  booktitle={International Conference on Learning Representations},
  year = {2021}
}

@inproceedings{liu2021infinite,
  title={Infinite nature: Perpetual view generation of natural scenes from a single image},
  author={Liu, Andrew and Tucker, Richard and Jampani, Varun and Makadia, Ameesh and Snavely, Noah and Kanazawa, Angjoo},
  booktitle={Proceedings of the IEEE/CVF International Conference on Computer Vision},
  pages={14458--14467},
  year={2021}
}

@inproceedings{ling2024dl3dv,
  title={Dl3dv-10k: A large-scale scene dataset for deep learning-based 3d vision},
  author={Ling, Lu and Sheng, Yichen and Tu, Zhi and Zhao, Wentian and Xin, Cheng and Wan, Kun and Yu, Lantao and Guo, Qianyu and Yu, Zixun and Lu, Yawen and others},
  booktitle={Proceedings of the IEEE/CVF Conference on Computer Vision and Pattern Recognition},
  pages={22160--22169},
  year={2024}
}

@inproceedings{yeshwanth2023scannet++,
  title={Scannet++: A high-fidelity dataset of 3d indoor scenes},
  author={Yeshwanth, Chandan and Liu, Yueh-Cheng and Nie{\ss}ner, Matthias and Dai, Angela},
  booktitle={Proceedings of the IEEE/CVF International Conference on Computer Vision},
  pages={12--22},
  year={2023}
}

@article{smart2024splatt3r,
  title={Splatt3r: Zero-shot gaussian splatting from uncalibrated image pairs},
  author={Smart, Brandon and Zheng, Chuanxia and Laina, Iro and Prisacariu, Victor Adrian},
  journal={arXiv preprint arXiv:2408.13912},
  year={2024}
}

@article{wang2004image,
  title={Image quality assessment: from error visibility to structural similarity},
  author={Wang, Zhou and Bovik, Alan C and Sheikh, Hamid R and Simoncelli, Eero P},
  journal={IEEE transactions on image processing},
  volume={13},
  number={4},
  pages={600--612},
  year={2004},
  publisher={IEEE}
}

@inproceedings{zhang2018unreasonable,
  title={The unreasonable effectiveness of deep features as a perceptual metric},
  author={Zhang, Richard and Isola, Phillip and Efros, Alexei A and Shechtman, Eli and Wang, Oliver},
  booktitle={Proceedings of the IEEE conference on computer vision and pattern recognition},
  pages={586--595},
  year={2018}
}

@inproceedings{schonberger2016structure,
  title={Structure-from-motion revisited},
  author={Schonberger, Johannes L and Frahm, Jan-Michael},
  booktitle={Proceedings of the IEEE conference on computer vision and pattern recognition},
  pages={4104--4113},
  year={2016}
}

@inproceedings{tang2024lgm,
  title={Lgm: Large multi-view gaussian model for high-resolution 3d content creation},
  author={Tang, Jiaxiang and Chen, Zhaoxi and Chen, Xiaokang and Wang, Tengfei and Zeng, Gang and Liu, Ziwei},
  booktitle={European Conference on Computer Vision},
  pages={1--18},
  year={2024},
  organization={Springer}
}

@inproceedings{zhang2024gs,
  title={Gs-lrm: Large reconstruction model for 3d gaussian splatting},
  author={Zhang, Kai and Bi, Sai and Tan, Hao and Xiangli, Yuanbo and Zhao, Nanxuan and Sunkavalli, Kalyan and Xu, Zexiang},
  booktitle={European Conference on Computer Vision},
  pages={1--19},
  year={2024},
  organization={Springer}
}

@inproceedings{jiang2024gaussianshader,
  title={Gaussianshader: 3d gaussian splatting with shading functions for reflective surfaces},
  author={Jiang, Yingwenqi and Tu, Jiadong and Liu, Yuan and Gao, Xifeng and Long, Xiaoxiao and Wang, Wenping and Ma, Yuexin},
  booktitle={Proceedings of the IEEE/CVF conference on computer vision and pattern recognition},
  pages={5322--5332},
  year={2024}
}

@inproceedings{NEURIPS2024_708e0d69,
 author = {Yang, Ziyi and Gao, Xinyu and Sun, Yang-Tian and Huang, Yi-Hua and Lyu, Xiaoyang and Zhou, Wen and Jiao, Shaohui and Qi, Xiaojuan and Jin, Xiaogang},
 booktitle = {Advances in Neural Information Processing Systems},
 editor = {A. Globerson and L. Mackey and D. Belgrave and A. Fan and U. Paquet and J. Tomczak and C. Zhang},
 pages = {61192--61216},
 publisher = {Curran Associates, Inc.},
 title = {Spec-Gaussian: Anisotropic View-Dependent Appearance for 3D Gaussian Splatting},
 volume = {37},
 year = {2024}
}

@inproceedings{liu2024mirrorgaussian,
  title={Mirrorgaussian: Reflecting 3d gaussians for reconstructing mirror reflections},
  author={Liu, Jiayue and Tang, Xiao and Cheng, Freeman and Yang, Roy and Li, Zhihao and Liu, Jianzhuang and Huang, Yi and Lin, Jiaqi and Liu, Shiyong and Wu, Xiaofei and others},
  booktitle={European Conference on Computer Vision},
  pages={377--393},
  year={2024},
  organization={Springer}
}

@article{jeong2026viewsplat,
  title={ViewSplat: View-Adaptive Dynamic Gaussian Splatting for Feed-Forward Synthesis},
  author={Jeong, Moonyeon and Min, Seunggi and Lee, Suhyeon and Seong, Hongje},
  journal={arXiv preprint arXiv:2603.25265},
  year={2026}
}

@article{xu2025resplat,
  title={ReSplat: Learning Recurrent Gaussian Splatting},
  author={Xu, Haofei and Barath, Daniel and Geiger, Andreas and Pollefeys, Marc},
  journal={arXiv preprint arXiv:2510.08575},
  year={2025}
}

@inproceedings{zhang2025flare,
  title={Flare: Feed-forward geometry, appearance and camera estimation from uncalibrated sparse views},
  author={Zhang, Shangzhan and Wang, Jianyuan and Xu, Yinghao and Xue, Nan and Rupprecht, Christian and Zhou, Xiaowei and Shen, Yujun and Wetzstein, Gordon},
  booktitle={2025 IEEE/CVF Conference on Computer Vision and Pattern Recognition (CVPR)},
  pages={21936--21947},
  year={2025},
  organization={IEEE}
}

@inproceedings{ye2026yonosplat,
  title={Yonosplat: You only need one model for feedforward 3d gaussian splatting},
  author={Ye, Botao and Chen, Boqi and Xu, Haofei and Barath, Daniel and Pollefeys, Marc},
  booktitle={International Conference on Learning Representations},
  volume={2026},
  pages={39852--39871},
  year={2026}
}

@article{oquab2023dinov2,
  title={Dinov2: Learning robust visual features without supervision},
  author={Oquab, Maxime and Darcet, Timoth{\'e}e and Moutakanni, Th{\'e}o and Vo, Huy and Szafraniec, Marc and Khalidov, Vasil and Fernandez, Pierre and Haziza, Daniel and Massa, Francisco and El-Nouby, Alaaeldin and others},
  journal={arXiv preprint arXiv:2304.07193},
  year={2023}
}

@inproceedings{ranftl2021vision,
  title={Vision transformers for dense prediction},
  author={Ranftl, Ren{\'e} and Bochkovskiy, Alexey and Koltun, Vladlen},
  booktitle={2021 IEEE/CVF International Conference on Computer Vision (ICCV)},
  pages={12159--12168},
  year={2021},
  organization={IEEE}
}

@article{hendrycks2016gaussian,
  title={Gaussian error linear units (gelus)},
  author={Hendrycks, Dan and Gimpel, Kevin},
  journal={arXiv preprint arXiv:1606.08415},
  year={2016}
}

@article{plucker1865xvii,
  title={Xvii. on a new geometry of space},
  author={Pl\"ucker, Julius},
  journal={Philosophical Transactions of the Royal Society of London},
  number={155},
  pages={725--791},
  year={1865},
  publisher={The Royal Society London}
}

@article{chen2024mvsplat360,
  title={Mvsplat360: Feed-forward 360 scene synthesis from sparse views},
  author={Chen, Yuedong and Zheng, Chuanxia and Xu, Haofei and Zhuang, Bohan and Vedaldi, Andrea and Cham, Tat-Jen and Cai, Jianfei},
  journal={Advances in Neural Information Processing Systems},
  volume={37},
  pages={107064--107086},
  year={2024}
}

@inproceedings{ICLR2025_9676c528,
 author = {Jin, Haian and Jiang, Hanwen and Tan, Hao and Zhang, Kai and Bi, Sai and Zhang, Tianyuan and Luan, Fujun and Snavely, Noah and Xu, Zexiang},
 booktitle = {International Conference on Learning Representations},
 editor = {Y. Yue and A. Garg and N. Peng and F. Sha and R. Yu},
 pages = {60001--60021},
 title = {LVSM: A Large View Synthesis Model with Minimal 3D Inductive Bias},
 volume = {2025},
 year = {2025}
}

@inproceedings{zhang2025transplat,
  title={Transplat: Generalizable 3d gaussian splatting from sparse multi-view images with transformers},
  author={Zhang, Chuanrui and Zou, Yingshuang and Li, Zhuoling and Yi, Minmin and Wang, Haoqian},
  booktitle={Proceedings of the AAAI Conference on Artificial Intelligence},
  volume={39},
  number={9},
  pages={9869--9877},
  year={2025}
}

@inproceedings{loshchilov2019iclr-decoupled,
  title     = {{Decoupled Weight Decay Regularization}},
  author    = {Loshchilov, Ilya and Hutter, Frank},
  booktitle = {International Conference on Learning Representations},
  year      = {2019}
}
\bibliographystyle{iclr2026_conference}
\newpage
\end{document}